# Strategy Selection in Influence Diagrams using Imprecise Probabilities

**Cassio P. de Campos**
Electrical, Computer and Systems Eng. Dept.
Rensselaer Polytechnic Institute
Troy, NY, USA
decamc@rpi.edu

**Qiang Ji**
Electrical, Computer and Systems Eng. Dept.
Rensselaer Polytechnic Institute
Troy, NY, USA
jiq@rpi.edu

## Abstract

This paper describes a new algorithm to solve the decision making problem in Influence Diagrams based on algorithms for credal networks. Decision nodes are associated to imprecise probability distributions and a reformulation is introduced that finds the global maximum strategy with respect to the expected utility. We work with Limited Memory Influence Diagrams, which generalize most Influence Diagram proposals and handle simultaneous decisions. Besides the global optimum method, we explore an anytime approximate solution with a guaranteed maximum error and show that imprecise probabilities are handled in a straightforward way. Complexity issues and experiments with random diagrams and an effects-based military planning problem are discussed.

## 1 INTRODUCTION

An influence diagram is a graphical model for decision making under uncertainty [13]. It is composed by a directed graph where utility nodes are associated to profits and costs of actions, chance nodes represent uncertainties and dependencies in the domain and decision nodes represent actions to be taken. Given an influence diagram, a strategy defines which decision to take at each node, given the information available at that moment. Each strategy has a corresponding expected utility. One of the most important problems in influence diagrams is *strategy selection*, where we need to find the strategy with maximum expected utility. A simple approach is to evaluate each possible strategy and compare their expected utilities. However, the number of strategies grows exponentially in the number of decision to be taken.

In this paper, we propose a new idea to find the best strategy based on a reformulation of the problem as an inference in a credal network [4]. We show through experiments that this approach can handle small and medium diagrams exactly, and provides an anytime approximation in case we stop the process early. Our idea works with a very general class of influence diagrams, named *Limited Memory Influence Diagrams* (LIMIDs) [15]. *Limited Memory* means that the assumption of *no-forgetting* usually employed in Influence Diagrams (that is, values of observed variables and decisions that have been taken are remembered at all later times) is relaxed. This class of diagrams is interesting because most other influence diagram proposals can be efficiently converted into LIMIDs.

To solve strategy selection, many approaches work on special cases of influence diagrams, exploiting their characteristics to improve performance. In many cases, it is assumed that there is an ordering on which the decisions are to be taken and the no-forgetting rule, so as previous decisions are assumed to be known in the moment of the current decision [14, 18, 19, 20, 21]. The ordering of decision nodes is exploited to evaluate the optimal strategy. There are also proposals in the class of simultaneous influence diagrams, where decisions are assumed to have no antecedents. This assumption reduces the number of possible strategies and allows for factorization ideas [22]. LIMIDs do not have assumptions about no-forgetting and ordering for decisions, even though it is possible to convert diagrams that have such assumptions into LIMIDs.

In order to test our method, we generate a data set of random influence diagrams. Empirical results indicate that the accuracy of our method is better than other approaches'. We also apply our idea to solve an Effects-based operations (EBO) military planning. The EBO approach seeks for a campaign objective by considering direct, indirect and cascading effects of military, diplomatic, psychological and economic actions [6, 11]. We use an influence diagram to model an EBO hypothetical problem.

Section 2 introduces our notation for influence diagrams and the problem of strategy selection. Section 3 describes the framework of credal networks and the inference problem on such networks. Section 4 presents how we solve strategy selection through a reformulation of the problem as an inference in credal networks. Section 5 presents some experiments, including the EBO military planning problem, and finally Section 6 concludes the paper and indicates future work.

## 2 INFLUENCE DIAGRAMS

A Limited Memory Influence Diagram $\mathcal{I}$ is composed by a directed acyclic graph $(\mathcal{V}, E)$ where nodes are partitioned in three types: chance, decision and utility nodes. Let $\mathcal{C}$, $\mathcal{D}$ and $\mathcal{U}$ be the set of chance, decision and utility nodes, respectively, and let $\mathcal{X} = \mathcal{C} \cup \mathcal{D}$. Links of $E$ characterize dependencies among nodes. Explicitly, links toward a chance node indicate probabilistic dependence of the node on its parents; links toward a decision node indicate which information is available to take such decision, and links toward utility nodes represent that an utility for those parents is to be considered (utility nodes may not have children). Associated to each node, there are some parameters:

1. A *chance node* has an associated categorical random variable $C$ with finite domain $\Omega_C$ and conditional probability distributions $p(C|\pi_j(C))$, for each configuration $\pi_j(C)$ of its parents $\pi(C)$ in the graph. $j$ is used to indicate a configuration of the parents of $C$, that is, $\pi_j(C) \in \Omega_{\pi(C)}$, where the notation $\Omega_{\mathcal{V}'} = \times_{V \in \mathcal{V}'} \Omega_V$, for any $\mathcal{V}' \subseteq \mathcal{V}$.

2. A *decision node* $D$ is associated to a finite set of mutually exclusive alternatives $\Omega_D$. Parents of $D$ describe the information that is available at the moment on which decision $D$ has to be taken.

3. An *utility node* $U$ is associated to a rational function $f_U : \Omega_{\pi(U)} \to \mathcal{Q}$. The value corresponding to a parent configuration is the profit (cost is viewed as negative profit) of such parent configuration. Utility nodes have no children.

A simple example is depicted in Figure 1. Decision nodes are represented by rectangles, chance nodes by ellipses and utility nodes by diamonds. *do_ground_attack* has an associated cost, which is depicted by the corresponding utility node. The same is modeled for *bomb_bridge*. The goal is to achieve *territory_occupation*, which also has an utility (the profit of the goal). *ground_attack* and *bridge_condition* represent the uncertain outcomes of the corresponding actions. Note that there is no known ordering on which decisions must be taken. Although decision nodes have no parents in this example, there is no such restriction.

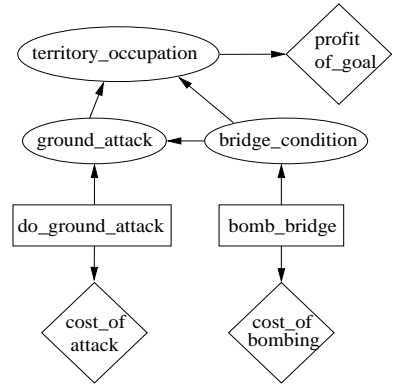

Figure 1: Simple Influence Diagram example.

A *policy* $\delta_D$ for the decision node $D$ is a function $\delta_D : \Omega_{D \cup \pi(D)} \to [0,1]$ defined for each alternative of $D$ and each configuration of $\pi(D)$ such that, for each $\pi_j(D) \in \Omega_{\pi(D)}$ we have $\sum_{d \in \Omega_D} \delta_D(d, \pi_j(D)) = 1$. A *pure policy* is a policy such that its image is integer ($\delta_D : \Omega_{D \cup \pi(D)} \to \{0,1\}$), and thus specifies with certainty which action (alternative of $D$) is taken for each parent configuration (in a pure policy, only one $\delta_D(d, \pi_j(D))$ for each $\pi_j(D)$ will be non-zero as they sum 1). A *strategy* $\Delta$ is a set of policies $\{\delta_D : D \in \mathcal{D}\}$, one for each decision node of the diagram. A *pure strategy* is composed only by pure policies.

The expected utility $EU(\Delta)$ of a strategy $\Delta$ is evaluated through the following equation:

$$\sum_{\mathbf{x} \in \Omega_{\mathcal{X}}} \left( \prod_C p(x_C | \pi_j(C)) \prod_D \delta_D(x_D) \sum_U f_U(\pi_{j'}(U)) \right), \quad (1)$$

where $x_C$, $\pi_j(C)$, $x_D$ and $\pi_{j'}(U)$ are respectively the projections of $\mathbf{x}$ in $\Omega_C$, $\Omega_{\pi(C)}$, $\Omega_{D \cup \pi(D)}$ and $\Omega_{\pi(U)}$. This equation means that, given a strategy, its expected utility is the sum of the utility values weighted by the probability of each diagram configuration (for all configurations). The maximum expected utility is obtained over all possible strategies:

$$MEU = \max_{\Delta} EU(\Delta).$$

The problem of *strategy selection* is to obtain the strategy that maximizes its expected utility, that is, $\operatorname{argmax} \max_{\Delta} EU(\Delta)$.

## 3 CREDAL NETWORKS

We need some concepts of credal networks before presenting the reformulation to solve strategy selection. A convex set of probability distributions is called a

*credal set* [4]. A credal set for $X$ is denoted by $K(X)$; we assume that every random variable is categorical and that every credal set has a finite number of vertices. Given a credal set $K(X)$ and an event $A$, the *upper* and *lower* probability of $A$ are respectively $\max_{p(X) \in K(X)} p(A)$ and $\min_{p(X) \in K(X)} p(A)$. A conditional credal set is a set of conditional distributions, obtained by applying Bayes rule to each distribution in a credal set of joint distributions.

A (separately specified) *credal network* $N = (G, \mathbb{X}, \mathbb{K})$ is composed by a directed acyclic graph $G = (V, E)$ where each node of $V$ is associated with a random variable $X_i \in \mathbb{X}$ and with a collection of conditional credal sets $K(X_i|\pi(X_i)) \in \mathbb{K}$, where $\pi(X_i)$ denotes the parents of $X_i$ in the graph. Note that we have a conditional credal set related to $X_i$ for each configuration $\pi_j(X_i) \in \Omega_{\pi(X_i)}$. A root node is associated with a single marginal credal set. We take that in a credal network every random variable is independent of its non-descendants non-parents given its parents; this is the *Markov condition* on the network. In this paper we adopt the concept of *strong independence*[1]: two random variables $X_i$ and $X_j$ are strongly independent when every extreme point of $K(X_i, X_j)$ satisfies standard stochastic independence of $X_i$ and $X_j$ (that is, $p(X_i|X_j) = p(X_i)$ and $p(X_j|X_i) = p(X_j)$) [4]. Strong independence is the most commonly adopted concept of independence for credal sets, probably due to its connection with standard stochastic independence.

Given a credal network, its *extension* is any joint credal set that satisfies all constraints encoded in the network. The *strong extension* $\mathcal{K}$ of a credal network is the largest joint credal set such that every variable is strongly independent of its non-descendants non-parents given its parents. The strong extension of a credal network is the joint credal set that contains every possible combination of vertices for all credal sets in the network [5]; that is, each vertex of a strong extension factorizes as follows:

$$p(X_1, \ldots, X_n) = \prod_i p(X_i|\pi(X_i)). \qquad (2)$$

Thus, a credal network can be viewed as a representation for a set of Bayesian networks with distinct parameters but sharing the same graph.

### 3.1 INFERENCE

A *marginal inference* in a credal network is the computation of upper (or lower) probabilities in an extension of the network. If $X_q$ is a *query* variable, then a marginal inference is the computation of tight bounds

---

[1]We note that other concepts of independence are found in the literature [3, 10].

---

for $p(x_q)$ for one or more categories $x_q$ of $X_q$. For inferences in strong extensions, it is known that distributions that maximize $p(x_q)$ belong to the set of vertices of the extension [12]. So, an inference can be produced by combinatorial optimization, as we must find a vertex for each local credal set $K(X_i|\pi(X_i))$ so that Expression (2) leads to a maximum of $p(x_q)$. In general, inference offers tremendous computational challenges, and exact inference algorithms based on enumeration of all potential vertices face serious difficulties [4].

A different way to solve the problem is to recognize that an upper (or lower) value for $p(x_q)$ may be obtained by the optimization of a multilinear polynomial over probability values, subject to constraints. This idea is discussed in the literature and different methods to reformulate the inference problem were proposed [7, 9]. Empirical results suggest that this is the most effective way for exact inferences. In the next section, we describe an idea based on bilinear programming [9] to perform inferences in credal networks and show how it can be employed to solve the strategy selection problem of influence diagrams.

## 4 STRATEGY SELECTION AS A CREDAL NET INFERENCE

Suppose we want to find the strategy $\Delta_{opt}$ that maximizes the expected utility in an influence diagram $\mathcal{I}$, that is, $\Delta_{opt} = \mathrm{argmax}\, MEU$. Let $\underline{f}$ and $\overline{f}$ be the minimum and maximum utility values specified in the diagram for all possible utility nodes and parent configurations, that is,

$$\underline{f} = \min_{U, \pi_j(U)} f_U(\pi_j(U)), \qquad \overline{f} = \max_{U, \pi_j(U)} f_U(\pi_j(U)).$$

We create an identical influence diagram $\mathcal{I}'$ except that the utility function $f'_U$ (for each node $U$) is defined as

$$\forall \pi_j(U) \quad f'_U(\pi_j(U)) = \frac{f_U(\pi_j(U)) - \underline{f}}{\overline{f} - \underline{f}}.$$

The denominator is positive because $\underline{f} < \overline{f}$ (if $\underline{f} = \overline{f}$, then the influence diagram is trivial as all utility values are equal). We note that this transformation is similar to that proposed by Cooper [2]. It is not hard to see that $\mathrm{argmax}\, MEU = \mathrm{argmax}\, MEU'$ (just take the terms out of summations in Equation (1)), and

$$\max_\Delta EU'(\Delta) = \frac{\max_\Delta EU(\Delta) - |\mathcal{U}|\underline{f}}{\overline{f} - \underline{f}}.$$

This implies that strategy selection in $\mathcal{I}$ is the same as strategy selection in $\mathcal{I}'$. Now, we translate the selection problem of $\mathcal{I}'$ to a credal network inference. Suppose we define a credal network with a similar graph as $\mathcal{I}'$ such that:

- Chance nodes are directly translated as nodes of the credal network (parents are the same as in $\mathcal{I}'$).

- Utility nodes are translated to binary random nodes. Let $U$ be an utility node with function $f_U$. In the credal network, $U$ becomes a binary node (with the same parents as before) and categories $u$ and $\neg u$ such that: $p(u|\pi_j(U)) = f_U(\pi_j(U))$ and $p(\neg u|\pi_j(U)) = 1 - p(u|\pi_j(U))$ [2].

- Decision nodes are translated to probabilistic nodes with imprecise distributions such that policies become probability distributions (in fact, according to our definition of policy, they are already greater than zero and sum 1). Thus, $p(d|\pi_j(D)) = \delta_D(d, \pi_j(D))$ for all $d$ and $\pi_j(D)$. Note that $p(D|\pi_j(D))$, for each $\pi_j(D)$, is a distribution with unknown probability values (this interpretation of decision nodes as imprecise probability nodes is discussed by Antonucci and Zaffalon, see e.g. [1]).

Using this credal network formulation, the expected utility of a strategy $\Delta$ can be written as

$$EU'(\Delta) = \sum_{\mathbf{x} \in \Omega_\mathcal{X}} \left( \prod_X p_\Delta(x|\pi_j(X)) \sum_U p(u|\pi_{j'}(U)) \right),$$

where $x$, $\pi_j(X)$ and $\pi_{j'}(U)$ are projections of $\mathbf{x}$ into the corresponding domains, $X$ ranges on all nodes corresponding to chance and decision nodes of the influence diagram, and $p_\Delta$ represents the distribution induced by the strategy $\Delta$, that is, when the strategy is chosen, $p_\Delta$ is a known probability distribution.

With some simple manipulations, we have:

$$EU'(\Delta) = \sum_{\mathbf{x} \in \Omega_X} \left( p_\Delta(\mathbf{x}) \sum_U p(u|\pi_{j'}(U)) \right),$$

$$EU'(\Delta) = \sum_{\mathbf{x} \in \Omega_X} \left( \sum_U p(u|\pi_{j'}(U)) p_\Delta(\mathbf{x}) \right),$$

$$EU'(\Delta) = \sum_U \sum_{\mathbf{x} \in \Omega_X} p_\Delta(u, \mathbf{x}) = \sum_U p_\Delta(u),$$

and then

$$MEU' = \max_\Delta \sum_U p_\Delta(u) = \max_{p \in \mathcal{K}} \sum_U p(u),$$

where $p \in \mathcal{K}$ means that we select a distribution $p$ in the extension of the credal network. In fact the only places $p$ may vary are related to the imprecise probabilities of the former decision nodes. When we select $p$, we get a precise distribution that has a corresponding strategy $\Delta$. So, we have a credal network and need to find a distribution $p$ that maximizes the sum of marginal probabilities of the $U$ nodes.

## 4.1 INFERENCE AS AN OPTIMIZATION PROBLEM

The sum of marginal inferences in the credal network can be formulated as a multilinear programming problem. The goal is to maximize the expression

$$\sum_U p(u) = \sum_U \sum_{\mathbf{x} \in \Omega_\mathcal{X}} \left( p(u|\pi_{j'}(U)) \prod_X p(x|\pi_j(X)) \right), \quad (3)$$

where $x$, $\pi_{j'}(U)$ and $\pi_j(X)$ are the projections of $\mathbf{x}$ in the corresponding domains, and where some distributions $p(X|\pi_j(X))$ are precisely known and others are imprecise. In this formulation we must deal with a large number of multilinear terms. To avoid them, we briefly describe the bilinear transformation procedure proposed by de Campos and Cozman [9] to replace the large Expression (3) by simple bilinear expressions. We refer to [9] for additional details.

The idea is based on a *precedence ordering* of the network variables, which is an ordering where all ancestors of a given variable in the network's graph appear before it in the ordering. The bilinear transformation algorithm processes the network variables top-down: at each step some constraints are generated that define the relationship between the query and the current variable being processed. A variable may be processed only if all its ancestors have already been processed. The active nodes at each step form a path-decomposition of the network's graph.

To better explain the method, we take the example of Figure 1. For simplicity, assume that variables are binary[2] (with categories $b$ and $\neg b$) renamed as follows: *do_ground_attack* is $D_1$, *bomb_bridge* is $D_2$, *cost_of_attack* is $U_1$, *cost_of_bombing* is $U_2$, *ground_attack* is $C_1$, *bridge_condition* is $C_2$, *territory_occupation* is $C_3$, and finally *profit_of_goal* is $U_3$.

After the translation of the utility functions into probability distributions and the replacement of decision nodes by nodes with imprecise probabilities (as previously described), we have a credal network and need to maximize the sum of the marginal probabilities of the $U$ nodes. In fact this is an extension of the standard query in a credal network, because we have a summation instead of a single probability to maximize. So the objective function is max $p(u_1) + p(u_2) + p(u_3)$ (there are three utility nodes in the example) subject to constraints that define each marginal probability $p(u_1)$, $p(u_2)$ and $p(u_3)$. To create these constraints, we run a symbolic inference based on the precedence ordering for each of the marginal probabilities. The constraints for $p(u_1)$ and $p(u_2)$ are very simple: $p(u_1) = p(u_1|d_1)p(d_1) + p(u_1|\neg d_1)p(\neg d_1)$ and $p(u_2) = p(u_2|d_2)p(d_2) + p(u_2|\neg d_2)p(\neg d_2)$, because they only depend on one other variable. Note that $p(d_1)$, $p(\neg d_1)$, $p(d_2)$, and $p(\neg d_2)$ that appear in these constraints are unknown and thus become optimization variables in the bilinear problem.

---

[2]The method works on non-binary variables as well. The assumption is made here for ease of expose.

To write the constraints for $p(u_3)$, we need to choose a precedence ordering. We will use the ordering $D_2, C_2, D_1, C_1, C_3, U_3$ (variables $U_1$ and $U_2$ do not appear in the order as they are not relevant to evaluate the marginal $p(u_3)$). Hence, the first variable to be processed is $D_2$. We write a constraint that relates the query $u_3$ and probabilities $p(D_2)$ (which are defined in the network specification):

$$p(u_3) = \sum_{d \in \{d_2, \neg d_2\}} p(d) \cdot p(u_3|d).$$

$D_2$ now appears in the conditional part of $p(u_3|d)$, which may be viewed as an artificial term in the optimization, as it does not appear in the network. Because of that, we must create constraints to define $p(u_3|d)$ in terms of network parameters (for all categories $d \in D_2$). According to our chosen ordering, the current variable to be processed is $C_2$. Thus,

$$p(u_3|d_2) = \sum_{c \in \{c_2, \neg c_2\}} p(c|d_2) \cdot p(u_3|c),$$

$$p(u_3|\neg d_2) = \sum_{c \in \{c_2, \neg c_2\}} p(c|\neg d_2) \cdot p(u_3|c).$$

Note that $p(u_3|c) = p(u_3|c,d)$ (for any $d$), so we use the simpler. At this stage, our query is conditioned on $C_2$. Following the same idea, we process $D_1$, obtaining

$$p(u_3|c_2) = \sum_{d \in \{d_1, \neg d_1\}} p(d) \cdot p(u_3|c_2, d),$$

$$p(u_3|\neg c_2) = \sum_{d \in \{d_1, \neg d_1\}} p(d) \cdot p(u_3|\neg c_2, d).$$

Now the current variable to be treated is $C_1$, and our query is conditioned on $C_2, D_1$, that is, we must define how to evaluate $p(u_3|C_2, D_1)$ for all configurations. Thus, for all $c \in \{c_2, \neg c_2\}$ and $d \in \{d_1, \neg d_1\}$:

$$p(u_3|c,d) = \sum_{c' \in \{c_1, \neg c_1\}} p(c'|c,d) \cdot p(u_3|c, c').$$

At this moment, $u_3$ is conditioned on $C_1, C_2$ in the artificial term $p(u_3|c, c')$ ($D_1$ is not present in the artificial term as $C_1, C_2$ separate $u_3$ from $D_1$). Now we process $C_3$: for all $c' \in \{c_1, \neg c_1\}$ and $c \in \{c_2, \neg c_2\}$

$$p(u_3|c, c') = \sum_{c'' \in \{c_3, \neg c_3\}} p(c''|c, c') \cdot p(u_3|c'').$$

Note that, as $p(u_3|c'')$ is specified in the network, we can stop. All artificial terms are related (through constraints) to parameters of the network. Besides all these constraints, we also include simplex constraints to ensure that probabilities sum 1.

Hence, we have a collection of linear and bilinear constraints on which non-linear programming can be employed [7]. It is also possible to use linear integer programming [9]. The steps to achieve a linear integer programming formulation are simple, because the only non-linear terms of the problem have the format $b \cdot t$, where $b \in \{0, 1\}$ and $t \in [0, 1]$. $b$ is an unknown probability value of the credal network (which is zero or one because the solution we look for lies on extreme points of credal sets [12]) and $t$ is a constant or an artificial term created in the procedure just described. To linearize the problem, $b \cdot t$ is replaced by an additional artificial optimization variable $y$ and the following constraints are inserted: $0 \leq y \leq b$ and $t - 1 + b \leq y \leq t$. After replacing all non-linear terms using this idea, the problem becomes a linear integer programming problem, where a solution is also a solution for the strategy selection in the initial influence diagram.

We emphasize that, as we are translating the strategy selection problem into a credal network inference, it is straightforward to use imprecise probabilities in the chance nodes of the influence diagram. Intervals or sets of probabilities may be used. The translation works in the same way, but the generated problem will have more imprecise probabilities to optimize.

The following theorem shows that, when reformulating the strategy selection problem as a modified credal network inference, we are not making use of "more effort" than necessary, that is, strategy selection has the same complexity as inference in credal networks.

**Theorem 1** *Let $\mathcal{I}$ be a LIMID and $k$ a rational. Deciding whether there is a strategy $\Delta$ such that MEU is greater than $k$ is NP-Complete when $\mathcal{I}$ has bounded induced width,[3] and $NP^{PP}$-Complete in general.*

*Proof sketch*: Pertinence for the bounded induced width case is achieved because (given a strategy) we can compute *MEU* and verify if it is greater than $k$ in polynomial time (using the reformulation and the sum of marginal queries, each marginal query takes polynomial time in a bounded induced width Bayesian network); in the general case, we can perform this verification using a PP oracle. Hardness for the bounded induced width case is obtained with the same reduc-tion as in [8] from the MAXSAT problem (replacing the credal nodes with decision nodes and introducing a single utility node). In the general case, the same reduction as in [17] from E-MAJSAT can be used (MAP nodes are replaced by decision nodes). □

---

[3]The maximum clique and the maximum degree in the moral graph are bounded by a logarithmic function in the size of the input needed to specify the problem, which for instance includes polytrees.

# 5 EXPERIMENTS

We conduct two experiments with the procedure. First, we use random generated influence diagrams to compare the solutions obtained by our procedure (which we call CR for *credal reformulation*) against the *Single Policy Updating* (SPU) of Lauritzen and Nilsson [15]. Later we work with a practical EBO military planning problem and compare the method against the factorization of Zhang and Ji [22].[4]

Concerning random influence diagrams, we have generated a data set based on the total number of nodes and the number of decision nodes. The configurations chosen are presented in the first two columns of Table 1. We have from 10 to 120 nodes, where 3 to 35 are decision nodes. The number of utility nodes is chosen equal to the number of decision nodes. Each line in Table 1 contains the average result for 30 random generated diagrams within that configuration. The third column of the table shows the approximate average number of distinct strategies in the diagrams that would need to be evaluated by a brute force method.

The three columns of the CR method show the time spent to solve the problem, the number of nodes evaluated in the branch-and-bound tree of the optimization procedure (which is significantly smaller than the total number of strategies in brute force) and the maximum error of the solution (all numbers are averages). After the reformulation, the CPLEX solver [16] is used, which includes a heuristic search before starting the branch-and-bound procedure. The evaluations of this heuristic search are not counted in the fifth column of Table 1. Note that the first five rows are separated from the last three because they strongly differ on the size of the search space (exact solutions were found only for the former). The maximum error of each solution is obtained straightforward from the relaxation of the linear integer problem. The last two columns of Table 1 show the time and maximum error of the SPU approximate procedure. Although very fast, the SPU procedure has worse accuracy than the "approximate" CR (solution was approximate in last three rows because we have imposed a time-limit of ten minutes for each run). Furthermore, SPU does not provide an upper bound for the best possible expected utility, as obtained by CR. Still, a possible improvement is to use SPU to provide an initial guess to the optimization.

## 5.1 EBO MILITARY PLANNING

In this section we describe the performance of our method in an hypothetical Effects-based Operations planning problem [11]. An influence diagram similar to the model described by Zhang and Ji [22] is employed. Its graph is shown in Figure 2. The goal is to win a war, which is represented by the *Hypothesis* node (on top of Figure 2). Just below there are the subgoals *Air_superiority*, *Territory_occupation*, and *Commander_surrender*, which are directly related to the main goal. There are eleven decision nodes (represented by rectangles): *destroy_C2* (C2 stands for *Command and Control*), *destroy_Radars*, *destroy_Communications*, *launch_air_strike*, *destroy_RD*, *destroy_storage*, *destroy_assembly*, *launch_ground_attack*, *launch_broadcasting*, *capture_bodyguard*, *use_special_force*. Just above decision nodes, we have chance nodes representing the outcomes of performing such actions (they indicate the workability of such systems), and below we have utility nodes (diamond-shaped nodes) describing the cost of each action. Furthermore, we have six chance nodes (in the center of the figure) indicating general workability of *IADS* (Integrated Air Defense System), *Air_force*, *Artillery*, *Ground_force*, *Morale* and *Commander_in_custody* with respect to enemy forces. The overall profit of winning is given by the node $U_H$, child of *Hypothesis*.

As this is an hypothetical example, we define utility functions and probability distributions as follows:

- Probability of *Hypothesis* is one given that all subgoals are achieved. If one of subgoals is not achieved, then the probability of *Hypothesis* is 60%; if two of them are not achieved, then the probability of success is 30%; if none of subgoals is achieved, then we certainly fail in the campaign.

- For the subgoals *Air_superiority*, *Territory_occupation*, and *Commander_surrender*, we define that the subgoal is accomplished with probability one when both children were achieved, 50% when only one child is achieved, and zero when none is achieved.

- For the probabilities of *IADS*, *Air_force*, *Artillery*, *Ground_force*, *Morale* and *Commander_in_custody*, we define a decrease of 50% for each unaccomplished child (with a minimum of zero, of course). Any node has probability zero if two or more of its children are not achieved.

- The outcomes of actions (chance nodes above decision nodes) have 90% of success. For exam-

---
[4]The factorization idea only works on simultaneous influence diagrams, so it was not used in the other test cases.

| Nodes | | Approx.# of | CR | | | SPU | |
|---|---|---|---|---|---|---|---|
| Total | Decision | Strategies | Time(sec) | Evals (B&B) | Max.Error(%) | Time(sec) | Max.Error(%) |
| 10 | 3 | $2^{17}$ | 0.66 | 5 | 0.000 | 0.10 | 0.740 |
| 20 | 6 | $2^{34}$ | 1.73 | 125 | 0.000 | 0.39 | 2.788 |
| 50 | 10 | $2^{51}$ | 30.42 | 4048 | 0.000 | 1.62 | 2.837 |
| 60 | 15 | $2^{52}$ | 29.77 | 2937 | 0.000 | 2.99 | 1.964 |
| 70 | 20 | $2^{54}$ | 125.06 | 7132 | 0.000 | 5.52 | 3.448 |
| 120 | 25 | $2^{102}$ | 254.80 | 15626 | 0.544 | 11.58 | 2.193 |
| 120 | 30 | $2^{116}$ | 403.13 | 5617 | 4.639 | 13.79 | 7.281 |
| 120 | 35 | $2^{120}$ | 578.99 | 9307 | 5.983 | 16.87 | 11.584 |

Table 1: Average results on 30 random influence diagrams of different sizes for the CR and SPU methods.

ple, *destroy_Radars* will have *EW/GCI_radars* destroyed with 90% of odds (EW/GCI means *Early Warning/Ground Control Interception*).

- The reward of achieving the main goal is 1000, while not achieving it costs 500.

- Costs of actions are as follows: *ground_attack* is 150, *use_special_force* is 100, *capture_bodyguard* is 80, *air_strike* is 50, and other actions cost 20 each.

For this problem, the best strategy found by SPU has expected utility of −55.2825, and suggests to take all action except *destroy_RD*, *destroy_storage*, *destroy_assembly* and *launch_ground_attack*. The global optimum strategy is found in less than 5 seconds with our method and has expected utility equal to 156.4051 (all actions are taken). This is much faster than the solution reported by [22] (around 45 seconds).

## 6 CONCLUSION

We discuss in this paper a new idea for strategy selection in Influence Diagrams. We work with the Limited Memory Influence Diagram, as it generalizes many of the influence diagram proposals. The main contribution is the reformulation of the problem as a credal network inference, which makes possible to find the global maximum strategy for small- and medium-sized influence diagrams. Experiments indicate that many instances can be treated exactly. As far as we know, no deep investigation of exact procedures for this class of diagrams has been conducted.

Because of the characteristics of our procedure, an anytime approximate solution with a maximum guaranteed error is available during computations. It is clear that large diagrams must be treated approximately. Nevertheless, in the conducted experiments, our method produced results that surpass existing algorithms. Although spending more time, many situations require a solution to be as good as possible, while time is a secondary issue. The ability of our approach to provide an upper bound for the result is also valuable, which is not available with the SPU method.

We also discuss the theoretical complexity of the problem, which is derived from the known properties of MAP problems in Bayesian networks and belief updating inferences in credal networks. The complexity results show that the proposed idea is not making use of a harder problem to solve a simpler one, as the complexity of strategy selection is the same as the complexity of inferences in credal networks.

Because strategy selection in influence diagrams and inferences in credal networks are related, improvements on algorithms of credal networks can be directly applied to influence diagram problems. The application of other approximate techniques based on credal networks seems a natural path for investigation. We also intend to explore other optimization criteria for influence diagrams with imprecise probabilities, besides expected utility. Proposals in the theory of imprecise probabilities might be applied to this setting.

## Acknowledgements

The work described in this paper is supported in part by the U.S. Army Research Office grant W911NF0610331.

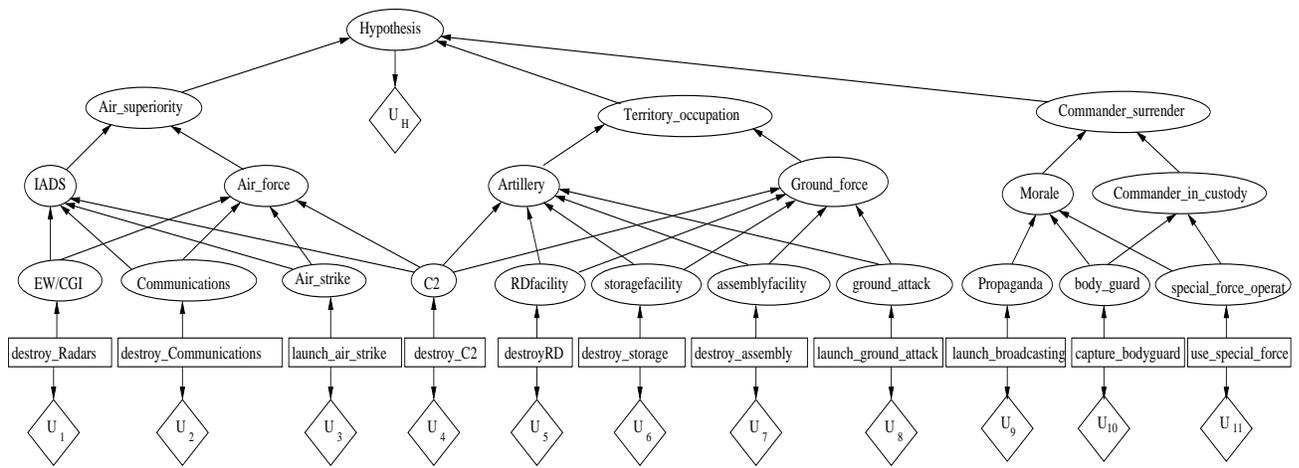

Figure 2: Influence Diagram for an hypothetical EBO-based planning problem.